\documentclass[sigconf]{acmart}

\AtBeginDocument{%
  }

\copyrightyear{2024}
\acmYear{2024}
\setcopyright{rightsretained}
\acmConference[HRI '24]{Proceedings of the 2024 ACM/IEEE International Conference on Human-Robot Interaction}{March 11--14, 2024}{Boulder, CO, USA}
\acmBooktitle{Proceedings of the 2024 ACM/IEEE International Conference on Human-Robot Interaction (HRI '24), March 11--14, 2024, Boulder, CO, USA}
\acmDOI{10.1145/3610977.3637471}
\acmISBN{979-8-4007-0322-5/24/03}

\usepackage[latin1]{inputenc}           
\usepackage{listings}
\usepackage{multirow}
\usepackage{cleveref}[2012/02/15]

\usepackage{ifthen}
\usepackage{tikz}
\usepackage{atbegshi,picture}
\newcommand{\myleftstd}{1in}

\newcommand\headertext{%
  \footnotesize Fares Abawi, Philipp Allgeuer, Di Fu, and Stefan Wermter. 2024. Wrapyfi: A
  Python Wrapper for Integrating Robots, Sensors, and Applications across
  Multiple Middleware. In Proceedings of the 2024 ACM/IEEE International
  Conference on Human-Robot Interaction (HRI '24), March 11-14, 2024, Boulder,
  CO, USA. \url{https://doi.org/10.1145/3610977.3637471}.
}

\newcommand{\setheader}{%
  \ifthenelse{\isodd{\thepage}}%
    {\newcommand{\myleftmargin}{\oddsidemargin+\myleftstd}}%
    {\newcommand{\myleftmargin}{\evensidemargin+\myleftstd}}%
    {{\oddsidemargin+\myleftstd}}%
    {{\evensidemargin+\myleftstd}}%
  \AtBeginShipoutNext{\AtBeginShipoutUpperLeft{%
    \put(\dimexpr\myleftmargin\relax,+1.2cm){\parbox{\textwidth}{\normalsize\sffamily\noindent\centering\hfill}}%
    \begin{tikzpicture}[remember picture,overlay]
      \node[anchor=north,yshift=-10pt] at (current page.north) {\parbox{\dimexpr\textwidth-\fboxsep-\fboxrule\relax}{\headertext}};
    \end{tikzpicture}%
  }}%
}

\let\citep\cite
\let\citet\cite
\let\citeyear\cite

\lstMakeShortInline[columns=fixed]|

\makeatletter
\gdef\@copyrightpermission{
  \begin{minipage}{0.3\columnwidth}
   \href{https://creativecommons.org/licenses/by/4.0/}{\includegraphics[width=0.90\textwidth]{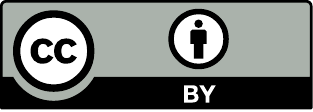}}
  \end{minipage}\hfill
  \begin{minipage}{0.7\columnwidth}
   \href{https://creativecommons.org/licenses/by/4.0/}{This work is licensed under a Creative Commons Attribution International 4.0 License.}
  \end{minipage}
  \vspace{5pt}
}
\makeatother

\begin{document}
\setheader
\title{Wrapyfi: A Python Wrapper for Integrating Robots, Sensors, and Applications across Multiple Middleware}

\author{Fares Abawi}
\email{fares.abawi@uni-hamburg.de}
\affiliation{%
  \institution{University of Hamburg}
  \streetaddress{Vogt-Koelln-Str. 30}
  \city{Hamburg}
  \country{Germany}
}

\author{Philipp Allgeuer}
\email{philipp.allgeuer@uni-hamburg.de}
\affiliation{%
  \institution{University of Hamburg}
  \streetaddress{Vogt-Koelln-Str. 30}
  \city{Hamburg}
  \country{Germany}
}

\author{Di Fu}
\email{di.fu@uni-hamburg.de}
\affiliation{%
  \institution{University of Hamburg}
  \streetaddress{Vogt-Koelln-Str. 30}
  \city{Hamburg}
  \country{Germany}
}

\author{Stefan Wermter}
\email{stefan.wermter@uni-hamburg.de}
\affiliation{%
  \institution{University of Hamburg}
  \streetaddress{Vogt-Koelln-Str. 30}
  \city{Hamburg}
  \country{Germany}
}


\begin{abstract}
  Message oriented and robotics middleware play an important role in facilitating robot control, abstracting complex functionality, and unifying communication patterns between sensors and devices. However, using multiple middleware frameworks presents a challenge in integrating different robots within a single system. To address this challenge, we present Wrapyfi, a Python wrapper supporting multiple message oriented and robotics middleware, including ZeroMQ, YARP, ROS, and ROS~2. Wrapyfi also provides plugins for exchanging deep learning framework data, without additional encoding or preprocessing steps. Using Wrapyfi eases the development of scripts that run on multiple machines, thereby enabling cross-platform communication and workload distribution. We finally present the three communication schemes that form the cornerstone of Wrapyfi's communication model, along with examples that demonstrate their applicability. \\ \textcolor{orange}{\url{http://software.knowledge-technology.info\#wrapyfi}}.
\end{abstract}




\begin{teaserfigure}
  \vspace{-0.9em}
  \centering
  \includegraphics[width=0.97\textwidth]{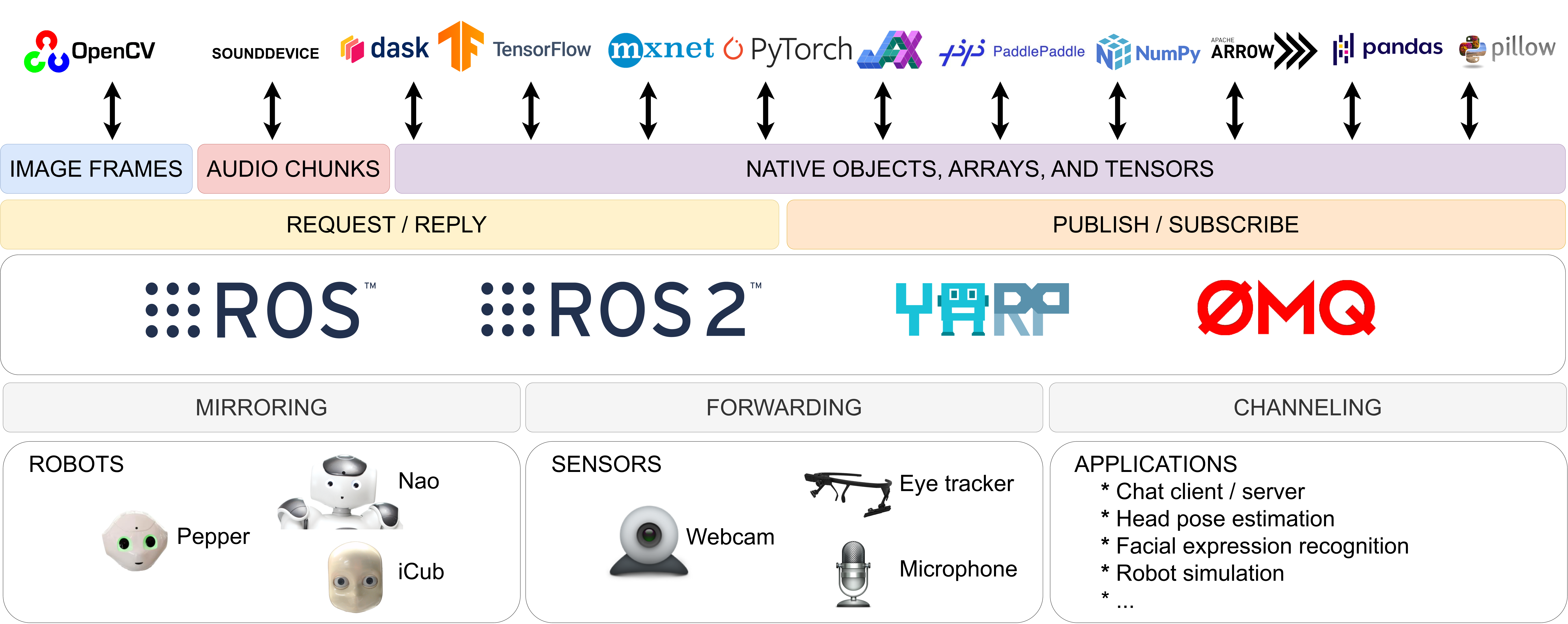}
  \vspace{-0.9em}
  \caption{Overview of the Wrapyfi framework. From top to bottom: 1) Data types are encoded or decoded depending on the transmission mode; 2) Encoded objects are prepared for transmission using the Request/Reply or Publish/Subscribe communication pattern; 3) Messages are transmitted through the selected middleware protocol; 4) Messages sequenced according to the communication scheme; 5) Messages exchanged between robots, applications, and sensors.$^\star$\\
  \tiny{\textcolor{gray}{$^\star$ The ``nine dots'' ROS and ROS 2 logos are trademarks of Open Source Robotics Foundation. TensorFlow, the TensorFlow logo, and any related marks are trademarks of Google Inc. The OpenCV logo is a trademark of https://opencv.org. The NumPy logo is used in accordance with the NumPy logo guidelines. The pandas logo is used in accordance with the brand and logo guidelines. PyTorch, the PyTorch logo and any related marks are trademarks of The Linux Foundation. The name ZeroMQ and the ``{\O}MQ'' logo are used in compliance with creative commons license Attribution-NoDerivatives 4.0 International (CC BY-ND 4.0). The logos for Dask, Apache MXNet, paddlepaddle, PIL (Pillow), JAX, and YARP are included with respect to their trademark policies; we acknowledge that these are subject to copyrights, trademarks, or registered trademarks of their respective holders. We do not claim ownership of these copyrights or trademarks. The use of these logos does not indicate endorsement by the trademark or copyright holders, nor does it suggest any affiliation or endorsement by the authors of this work.}}}
  \Description{Overview of the Wrapyfi framework, showing the supported frameworks, libraries, encoding and datatypes, middleware, and schemes of communication. The potential applications and interfaces are shown}
  \label{fig:intro_teaser}
\end{teaserfigure}


\maketitle

\section{Introduction}
Real-time robotic applications require exchanging multimodal data arriving from a variety of sensors. A framework that distributes sensory information across processes is necessary, especially for robot-robot and human-robot interaction~\citep{mohamed2021ros}. Multiprocess and multithread instances are used to parallelize independent methods. However, such parallelization approaches are limited to single machines and may not be sufficient for applications with a large number of sensors or computationally expensive processing methods. Eventually, this leads to performance bottlenecks on consumer-grade computers. Message oriented and robotics middleware, such as ZeroMQ~\citep{zeromq2013hintjens}, YARP~\citep{yarp2006metta}, ROS~\citep{ros2009quigley}, and ROS~2~\citep{ros22022macenski}, were developed to tackle such challenges. Middleware frameworks use communication protocols to exchange data and distribute operations across several machines and nodes~\citep{elkady2012middleware}. 

ROS~\citep{ros2009quigley} is a middleware commonly used in the robotics community. ROS provides control hardware interfaces, visualization tools, and communication models for many robotic platforms~\citep{abi2019osrros}. Its widespread use is a direct result of its early adoption of open source and the vast amount of robotic tools provided by its developers and contributors. However, ROS is scheduled for deprecation in favor of ROS~2~\citep{ros22022macenski}. Many robotic platforms and packages, nonetheless, have not been updated to support this transition yet. Although bridges were developed to enable communication between ROS, ROS~2, and WebSocket, integrating such bridges into working pipelines requires major modifications to the underlying code and its structure. This demands following certain naming conventions and limiting the message types supported, resulting in additional effort. Other middleware designed specifically for certain robotic platforms such as YARP~\citep{yarp2006metta} used by the iCub~\citep{metta2010icub} robot, provide interfaces for communicating with ROS~\cite{natale2016icub} as well. However, their usage dictates modifying scripts to accommodate specific message types. This poses a major hurdle for developers wanting to integrate different robots and middleware, as a result, restricting the cross-compatibility of their applications with existing systems.

To improve interoperability between different robotic platforms and reduce reliance on a particular middleware, we have developed the open source Wrapyfi\footnote{\label{wrapyfi_link}\url{https://github.com/fabawi/wrapyfi}} (illustrated in~\autoref{fig:intro_teaser}) framework, a Python wrapper supporting multiple middleware bindings. Wrapyfi is a simpler alternative to GenoM3~\citep{mallet2010genom3}. GenoM3 adopts a model-driven approach and uses templates to define the components and data exchanges across middleware. Since it is specifically developed for Python scripting, Wrapyfi eliminates the need for having to learn another language or to define templates, unlike GenoM3. REMS~\citep{rems2022tanaka} is a middleware built in Python with simplistic interfaces for educational purposes. Although REMS supports a large set of robots and simulation environments, it does not address interoperability between different middleware operating on them. 

Wrapyfi's decorator-based design integrates easily with existing workflows, prioritizing minimal modifications for improved multi-robot communication. Beyond robotic applications, its adaptability is observed in supporting message oriented middleware, facilitating communication with interfaces that do not necessarily require the additional packages and tools provided by robotics middleware. Deep learning frameworks like JAX~\citep{jax2018bradbury} and PyTorch~\citep{pytorch2019paszke}, support multi-machine parallelization mainly through remote procedure calls. The approaches adopted in distributing models and data differ greatly, including the communication patterns used and the orchestration of communication, having either a single or several controllers. By offering a standard approach for multiple frameworks, and supporting two of the most common communication patterns, namely publish-subscribe and request-reply---also known as the request-response or client-server pattern---Wrapyfi offers greater control over communication dynamics in comparison to each framework's parallelization protocol.

Open Neural Network Exchange(ONNX)~\citep{onnx2019bai} is a framework designed to standardize machine learning model representations, offering compatibility with a wide range of deep learning frameworks. However, using ONNX with any framework requires converting the model formats. In contrast, Wrapyfi does not impose such a constraint or bind developers to a specific protocol. Wrapyfi does not only allow for native Python object exchanges but also transports data structures such as arrays and tensors, which are relied upon in deep learning applications. This integration makes Wrapyfi a useful tool for developers, allowing them to take advantage of both robotics and deep learning ecosystems.

\section{Data types}
Wrapyfi employs a type-aware serialization method that automatically transforms the objects exchanged between script mirrors into a format compatible with the selected middleware. Wrapyfi supports the following data types:

\bigskip

\noindent \normalfont{\textbf{Native objects, arrays, and tensors.}
Wrapyfi allows for the transmission of a variety of data types used in Python. Prior to transmission, these data types are converted into JSON strings to ensure compatibility across different middleware platforms. Wrapyfi supports using NumPy~\citep{numpy2020harris} arrays and enables their sharing across mirrored scripts. Moreover, Wrapyfi offers a plugin interface that developers may use to customize the transmission of other types of objects. This feature allows encoding objects as strings, which can eventually be decoded back into their original structure. Wrapyfi comes with built-in plugins for exchanging Arrow~\citep{richardson2023arrow} vectors, pandas\footnote{pandas version 1 with NumPy as a backend}~\citep{reback2020pandas} data frames, and Pillow\footnote{\url{https://github.com/python-pillow/Pillow}} images. It also supports tensors from major deep learning frameworks such as TensorFlow~\citep{tensorflow2015abadi}, PyTorch~\citep{pytorch2019paszke}, MXNet~\citep{mxnet2015chen}, JAX~\citep{jax2018bradbury}, PaddlePaddle~\citep{paddlepaddle2019ma}, and Dask~\citep{dask2015rocklin}. These plugins make it possible to exchange data between different frameworks and to integrate deep learning models into robotic systems.
When specified, the tensors transmitted using Wrapyfi can be mapped to GPUs or CPUs different from the ones specified on a publishing script's end, allowing for the distribution of computationally demanding deep learning models.} 

\bigskip

\noindent \normalfont{\textbf{Images.} ROS, ROS~2, and YARP provide specialized message types for transmitting images. We use image messages to stream raw monochrome, RGB, and JPEG-encoded images. ZeroMQ does not provide such specialized message structures. Therefore, we make use of the multipart message structure to create an image interface, allowing us to standardize middleware behavior and transmit the image properties to a specified topic.}

\bigskip

\noindent \normalfont{\textbf{Audio chunks.} ROS and ROS~2 do not provide messages structured for audio transmission, so we create custom messages and services to transmit audio along with its properties. The number of audio channels transmitted can vary in size, as long as the audio chunk structure follows the python-sounddevice format\footnote{\url{https://github.com/spatialaudio/python-sounddevice}}. For YARP, we use the existing sound port and transmit the audio as a sequence. Whereas, for ZeroMQ, we transmit a string, encoding the auditory signal along with its properties as a single multipart message.}

\section{Communication schemes}

Wrapyfi manages script interactions using three communication schemes---Mirroring, Forwarding, and Channeling. Mirroring enables concurrent execution of multiple scripts with synchronized actions. Forwarding creates chains of methods to tunnel arguments and return values across different middleware configurations. Channeling allows for the broadcasting of multiple return values via one method, each using potentially different middleware. Each scheme addresses different challenges in distributed systems.
\begin{lstlisting}[frame=single, language=Python, caption={Decorated method registering the data type, middleware, topic, connection protocol, and blocking behavior. \lstinline{'$0'} passes the first argument (\lstinline{mware}) from the method to the decorator. Similarly, \lstinline{'$blocking'} passes the keyword argument.\\}, label=fig:intro_example, captionpos=t, basicstyle=\footnotesize, numbers=left, upquote=true, xleftmargin=4.0ex, xrightmargin=1.5ex, abovecaptionskip=5pt, belowcaptionskip=5pt,  showstringspaces=false]
class MirrorCls(MiddlewareCommunicator):
  @MiddlewareCommunicator.register('NativeObject', 
   '$0', 'MirrorCls', '/example/read_msg', 
   carrier='tcp', should_wait='$blocking')
  def read_msg(self, mware, msg='', blocking=True):
    msg_ip = input('type message:')
    obj = {'msg': msg, 'msg_ip': msg_ip}
    return obj,
\end{lstlisting}

\bigskip

The \lstinline{MiddlewareCommunicator} is a Wrapyfi class for establishing communication methods. It implements the \lstinline{register} decorator for setting the middleware types, topics, and various communication parameters. Each method set to publish, subscribe, request, or reply should be encapsulated with this decorator. \autoref{fig:intro_example} illustrates the use of the \lstinline{register} decorator to register a method for YARP middleware communication, specifying object type, middleware, name of the class, YARP port (topic), communication protocol, and whether the method should await a response, which results in blocking the subscribing method until the publisher transmits a message. The \lstinline{read_msg} method obtains user input from one process, allowing all other subscribing processes to acquire user input from a single process invocation.

\begin{lstlisting}[frame=single, language=Python, caption={Activating a method in \lstinline{'publish'} mode. When the method is called, its results are returned to the caller and transmitted to the listener.\\}, label=fig:activate_communication, captionpos=t, basicstyle=\footnotesize, numbers=left, upquote=true, xleftmargin=4.0ex, xrightmargin=1.5ex, abovecaptionskip=5pt, belowcaptionskip=5pt,  showstringspaces=false]
mirror = MirrorCls()
mirror.activate_communication(
    'read_msg', mode='publish')
\end{lstlisting}

\bigskip

In \autoref{fig:activate_communication}, setting the mode to \lstinline[upquote=true]{'publish'} triggers \lstinline{read_msg} upon method call, whereas \lstinline[upquote=true]{'listen'} returns the message received over the middleware. These modes enable the establishment of communication following the publish/subscribe pattern. Alternatively, setting the \lstinline[upquote=true]{activate_communication} mode to \lstinline[upquote=true]{'request'} or \lstinline[upquote=true]{'reply'} triggers the request/reply pattern.

\bigskip

\noindent \normalfont{\textbf{Mirroring.} Mirrors are multiple scripts running concurrently. The scripts share arguments and return values using a predefined communication pattern. The behaviors of all mirrored scripts are identical. However, their methods could either execute functionality in place or acquire their return values from another publisher. By calling \lstinline{read_msg} in \autoref{fig:intro_example} using a single publishing script, all subscribing mirrors receive the same return object when invoked as well. Regardless of the communication pattern or blocking behavior, all scripts follow the same pipeline with similar method returns.}

\bigskip

\noindent \normalfont{\textbf{Forwarding.} The forwarding scheme in Wrapyfi enables passing arguments to multiple methods, each with a different middleware setting. This forms a chain of methods, transferring arguments and return values across middleware and topics. Forwarding employs multiple scripts with unique functions, connected by \lstinline{register} decorators, making it suitable for creating multi-step processes with several scripts having partial component support. In \autoref{fig:forwarding_example}, we demonstrate data transmission between a system without ZeroMQ support and another without Yarp support, using an intermediary system that supports both. The first system dispatches the message using Yarp by invoking \lstinline{send_yarp}. The intermediary system then forwards it using ZeroMQ to \lstinline{send_zmq}. The final system, with Yarp disabled, receives the message via ZeroMQ by listening to \lstinline{send_zmq}. This scheme is needed when strict specifications are required regarding compatibility of software and middleware between systems, as in the case of robots.}

\begin{lstlisting}[frame=single, language=Python, caption={Demonstration of forwarding with two methods each using a different middleware.\\}, label=fig:forwarding_example, captionpos=t, basicstyle=\footnotesize, numbers=left, upquote=true, xleftmargin=4.0ex, xrightmargin=1.5ex, abovecaptionskip=5pt, belowcaptionskip=5pt,  showstringspaces=false]
class ForwardCls(MiddlewareCommunicator):
  @MiddlewareCommunicator.register('NativeObject', 
   'yarp', 'ForwardCls', '/example/native_yarp_msg', 
   carrier='mcast', should_wait=True)
  def send_yarp(self, msg):
    return msg,

  @MiddlewareCommunicator.register('NativeObject', 
   'zeromq', 'ForwardCls', '/example/native_zmq_msg', 
   carrier='tcp')
  def send_zmq(self, msg):
    return msg,
\end{lstlisting}

\bigskip

\noindent \normalfont{\textbf{Channeling.} In the channeling scheme, Wrapyfi enables broadcasting to multiple middleware by encapsulating a method with numerous decorators, each corresponding to a return value with its own data type and middleware. This is illustrated in \autoref{fig:channel_example}, where a method transmits three different data types over varied middleware, such as a Yarp native object message comprising a NumPy image and an audio chunk, a ROS image (OpenCV~\citep{opencv2000bradski} compatible), and a ZeroMQ audio chunk. This scheme supports the simultaneous reception of different data types. If the environment lacks support for a specified middleware, a \lstinline{None} type object is returned. Channeling is especially useful for handling multiple sensory inputs from different sources, allowing selective acquisition and disregard of unnecessary sensory input. This provides a balanced approach between mirroring and forwarding, altering the pipeline based on the returns received from the supported middleware.}

\begin{lstlisting}[frame=single, language=Python, caption={Demonstration of Channeling with one method reading multiple returns of different data types through multiple middleware.\\}, label=fig:channel_example, captionpos=t, basicstyle=\footnotesize, numbers=left, upquote=true, xleftmargin=4.0ex, xrightmargin=1.5ex, abovecaptionskip=5pt, belowcaptionskip=5pt,  showstringspaces=false]
class ChannelCls(MiddlewareCommunicator):
  @MiddlewareCommunicator.register('NativeObject', 
   'yarp', 'ChannelCls', '/example/native_yarp_msg', 
   carrier='mcast', should_wait=True)
  @MiddlewareCommunicator.register('Image', 
   'ros', 'ChannelCls', '/example/image_ros_msg', 
   carrier='tcp', width='$img_width', 
   height='$img_height', rgb=True, queue_size=10)
  @MiddlewareCommunicator.register('AudioChunk', 
   'zeromq', 'ChannelCls', '/example/audio_zmq_msg', 
   carrier='tcp', rate='$aud_rate', 
   chunk='$aud_chunk', channels='$aud_chann')
  def read_mulret_mulmware(self, 
   img_width=200, img_height=200, 
   aud_rate=44100, aud_chunk=8820, aud_chann=1):
    ros_img = np.random.randint(256, 
     size=(img_height, img_width, 3), dtype=np.uint8)
    zeromq_aud = (np.random.uniform(-1,1, aud_chunk), 
     aud_rate,)
    yarp_native = [ros_img, zeromq_aud]
    return yarp_native, ros_img, zeromq_aud
\end{lstlisting}

\section{Use Cases}

\begin{figure}[!htbp]
        \centering
        \includegraphics[width=0.4722\textwidth,trim={0 0 0 0.25in},clip]{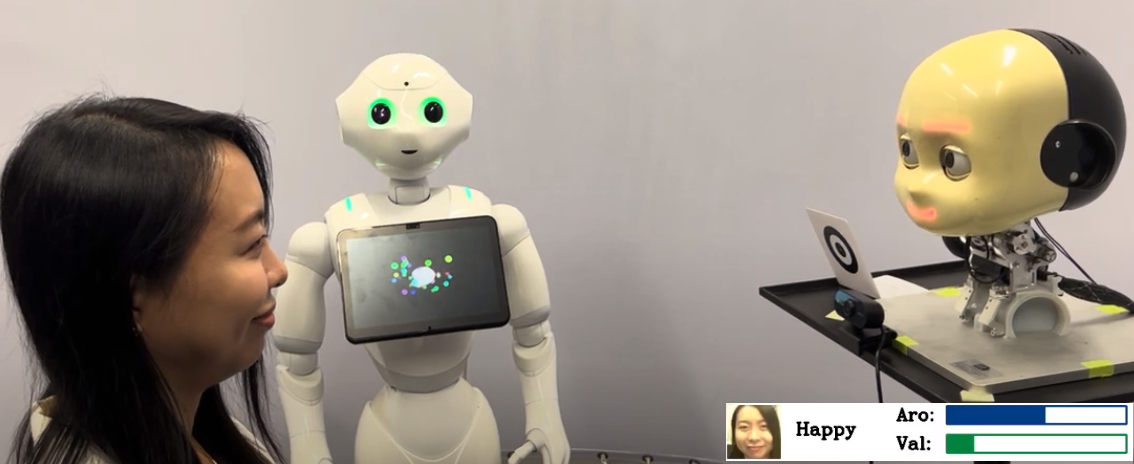}
        \vspace*{-3mm}
        \caption{Facial expression imitation on the Pepper and iCub.}
        \label{fig:imitation_multirobot}
    \end{figure}

\noindent \normalfont{\textbf{Facial expression imitation.}  Participants exhibit eight facial expressions while sitting in front of two robots, Pepper~\citep{pepper2014softbank} and iCub~\citep{metta2010icub} as depicted in~\autoref{fig:imitation_multirobot}. The robots then imitate the participants' expressions. Pepper represents emotions through color changes, while iCub displays robotic facial expressions. The forwarding scheme in Wrapyfi tunnels interactions between the different system components and middleware configurations, enabling the exchange of visual and facial expression data between the robots and the recognition model~\citep{siqueira2020efficient}. Forwarding manages image acquisition across robots and synchronizes the transfer of facial expressions to and from the model by sequentially invoking each robot's acquisition and action methods.}

\bigskip 

\begin{figure}[!htbp]
        \centering
        \includegraphics[width=0.4722\textwidth]{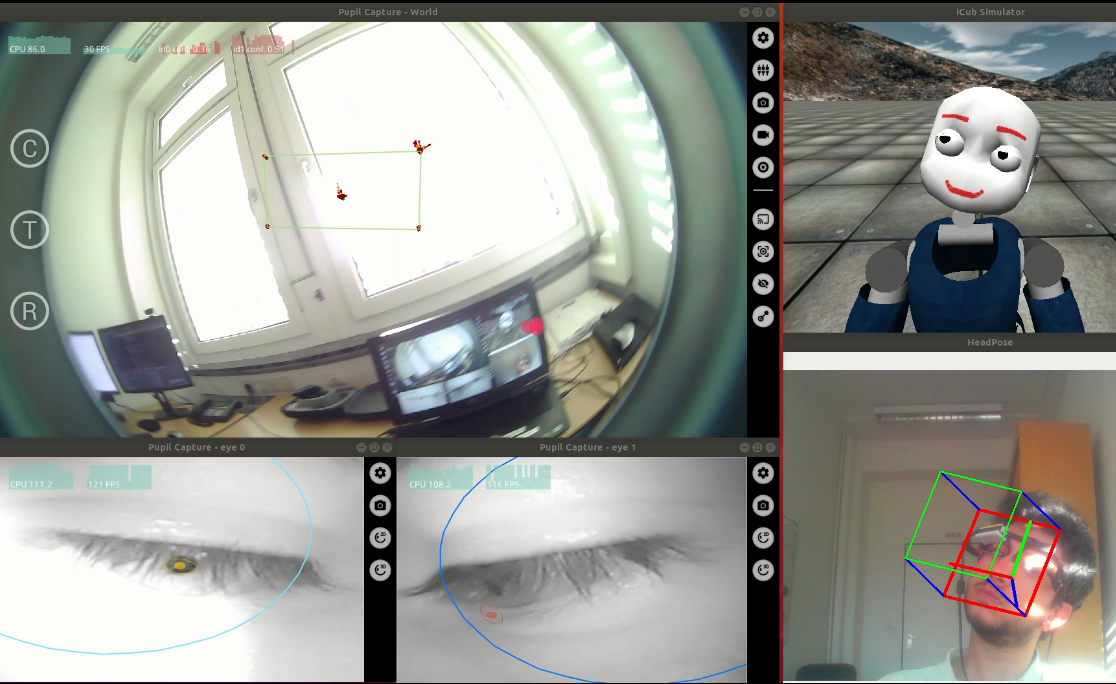}
        \vspace*{-2mm}
        \caption{Head and eye movement imitation using either an IMU-fitted eye tracker or a head pose estimation model.}
        \label{fig:imitation_multisensor}
    \end{figure}

\noindent \normalfont{\textbf{Head orientation imitation.} In this example, we imitate a participant's head orientation and eye movements on a simulated iCub~\citep{tikhanoff2008icubsim} as shown in~\autoref{fig:imitation_multisensor}. The input coordinates arrive either from a wearable eye tracker~\citep{kassner2014pupil} fitted with an IMU or a vision-based head pose estimation model~\citep{hempel2022sixdrepnet}. The channeling scheme allows switching between the input sources by specifying the return element propagated to the robot.}

\section{Code and Usage}
To install Wrapyfi, compatible middleware, and required interfaces:\\
\textcolor{orange}{\url{https://wrapyfi.readthedocs.io/}}. \\
We additionally provide instructions on running Wrapyfi examples:\\
\textcolor{orange}{\url{https://wrapyfi.readthedocs.io/en/latest/examples.html}}. \\
Tutorials detail the steps needed to run the Wrapyfi use case scripts:\\
\textcolor{orange}{\url{https://wrapyfi.readthedocs.io/en/latest/tutorials.html}}. \\
We also evaluate transmission latency of the Wrapyfi plugins:\\
\textcolor{orange}{\url{https://wrapyfi.readthedocs.io/en/latest/evaluation.html}}.

\section{Conclusions}
Wrapyfi is a framework that simplifies data transfer across different middleware platforms. Two of Wrapyfi's key strengths are the transmission of custom data types and support for multiple middleware. We introduced three communication schemes---mirroring, forwarding, and channeling---each serving a different set of applications. The framework currently supports two common communication patterns: publish-subscribe and request-reply. In future work, we plan to extend Wrapyfi to support more communication patterns that are available in some middleware platforms, such as actions in ROS~2, which are similar to asynchronous request-reply. We also aim to provide interfaces for custom messages and middleware-specific data types. Wrapyfi's modular design permits integrating further middleware, expanding the array of potential applications. 

\begin{acks}
The authors gratefully acknowledge partial support from the German Research Foundation DFG under project CML~(TRR~169).
\end{acks}





\bibliographystyle{ACM-Reference-Format}
\balance
\bibliography{submission}










\end{document}